\documentclass[letterpaper, 10 pt, conference]{ieeeconf}  

\IEEEoverridecommandlockouts                              
                                                          
\usepackage{graphicx}
\usepackage{cite}
\usepackage{gensymb}
\usepackage{siunitx}

\graphicspath{ {./images/} }

\title{\LARGE \bf
GelLink: A Compact Multi-phalanx Finger with Vision-based Tactile Sensing and Proprioception
}

\author{
    \authorblockN{Yuxiang Ma, Jialiang (Alan) Zhao, Edward Adelson}
        \authorblockA{Massachusetts Institute of Technology\\
    {\tt\small yxma20@mit.edu, alanzhao@mit.edu, adelson@csail.mit.edu}} 
}

\begin{document}

\maketitle
\thispagestyle{empty}
\pagestyle{empty}

\begin{abstract}
Compared to fully-actuated robotic end-effectors, underactuated ones are generally more adaptive, robust, and cost-effective. 
However, state estimation for underactuated hands is usually more challenging.
Vision-based tactile sensors, like Gelsight, can mitigate this issue by providing high-resolution tactile sensing and accurate proprioceptive sensing. 
As such, we present GelLink, a compact, underactuated, linkage-driven robotic finger with low-cost, high-resolution vision-based tactile sensing and proprioceptive sensing capabilities.
In order to reduce the amount of embedded hardware, i.e. the cameras and motors, we optimize the linkage transmission with a planar linkage mechanism simulator and develop a planar reflection simulator to simplify the tactile sensing hardware. 
As a result, GelLink only requires one motor to actuate the three phalanges, and one camera to capture tactile signals along the entire finger.
Overall, GelLink is a compact robotic finger that shows adaptability and robustness when performing grasping tasks. 
The integration of vision-based tactile sensors can significantly enhance the capabilities of underactuated fingers and potentially broaden their future usage.



\end{abstract}

\section{INTRODUCTION}
In recent years, a lot of dexterous and anthropomorphic hands have been developed to replicate human hands. These dexterous designs usually include more than ten degrees of freedom (DOFs) and an equivalent number of actuators \cite{xiong2016design, hundhausen2020soft, puhlmann2022rbo, bdhand, kim2021integrated}. Compared to parallel-jaw grippers with only one DOF, these more complex hands usually achieve better performance in high-precision manipulation tasks or tasks that involve interacting with complicated geometries, such as tool-use and in-hand manipulation \cite{xiong2016design, hundhausen2020soft}. 
However, it is challenging to balance functionality and complexity when designing a finger because arranging a large number of joints and actuators in a robot hand is non-trivial and costly \cite{puhlmann2022rbo, kim2021integrated}.
In addition, having a large number of DOFs also leads to difficulty in motion planning and control \cite{gosselin2008anthropomorphic, catalano2014adaptive}. 

Underactuated grippers have become more wide-spread because they are light-weight and easy to control.
Because underactuated fingers have more DOFs than actuators \cite{birglen2007underactuated}, they are capable of mimicking the natural conforming motion or ``adaptation'' of human fingers. 
This ``adaptation'' enables them to grasp a wide array of objects stably without the need for complicated control policies and motion planning. 

However, many underactuated fingers exhibit reduced precision and limited control due to this lack of actuation. For tasks that require high precision and fine control, developing effective control strategies for underactuated fingers can be complicated, and it may require advanced machine learning techniques or sophisticated control frameworks. 

Sensory feedback, such as proprioception and tactile sensing, can potentially augment the functionality of underactuated fingers. Proprioception, also commonly referred to as the perception of finger configuration, can be used to improve control precision \cite{she2020exoskeleton}. Meanwhile, tactile sensing can perceive rich contact information, such as contact geometry, contact force or torque, slip, and vibration, which can be further used to implement feedback controllers and enhance their performance in manipulation tasks. 

\begin{figure}[h!]
    \centering
    \includegraphics[width=0.48\textwidth]{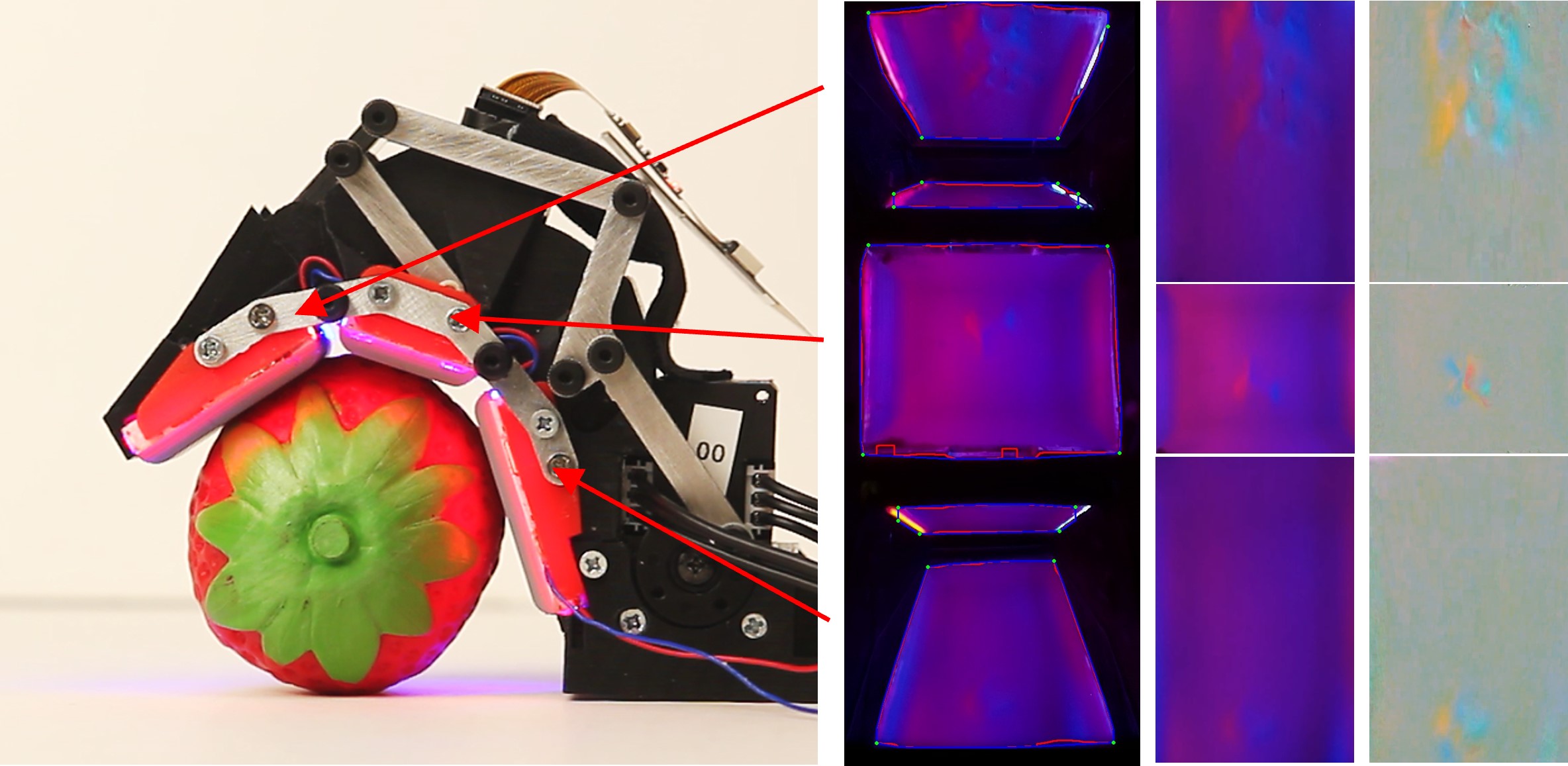}
    \caption{From left to right: GelLink touching a plastic strawberry, the corresponding raw tactile images, unwarped images, and difference images}
    \label{fig:teaser}
\end{figure}

This work proposes GelLink, an underactuated robotic finger incorporated with high-resolution and low-cost camera-based tactile and proprioceptive sensing. 
Furthermore, we introduce a multi-link mirror-based mechanism so that one camera is able to capture the entire sensing area at any configuration.
The main design goal of GelLink is to limit the amount of hardware needed and still maintain the functionality and versatility of a multi-phalanx finger with embedded tactile sensing.
The final design has three phalanges and two DOFs, controlled by only one motor and visualized by only one camera.
The fingers can grasp objects of various shapes and materials while providing rich geometry and textural information of grapsed objects. 





\section{RELATED WORK}

\subsection{Underactuated fingers integrated with tactile sensing}

Underactuated mechanisms have fewer DOFs compared with fully actuated ones, making them suitable for lightweight and compact robotic end-effector designs. 
Researchers have combined tactile sensors with underactuated fingers to augment their performance in manipulation tasks.
Most work that involves both underactuation and tactile sensing uses pressure or force sensors, which convert changes in pressure or force to changes in electric properties, such as capacitance, resistance, or voltage. These sensors are usually fast and small, and they are suitable for robotic integration.

Spiers \textit{et al.} implemented feature extraction with a two-phalanx finger that has eight embedded barometric pressure sensors \cite{spiers2016single}.  
Yoon \textit{et al.} proposed a tendon-driven robotic finger design that incorporated stretchable tactile sensors based on the piezoresistive effect of liquid metal \cite{yoon2022elongatable}. 
Using the sensor information, the gripper was able to grasp an inflated balloon and maintain a desired contact pressure with force feedback control. 
Lu \textit{et al.} combined an underactuated linkage finger with arrays of biomimetic tactile sensors, which measure both normal and shear forces \cite{lu2022gtac}. 
Tactile feedback facilitated the gripper to achieve grasping and simple in-hand manipulation.
Their gripper was able to grasp various objects and achieve in-hand manipulation with closed-loop control. 

Although many researchers have shown promising results of augmenting underactuated fingers with force sensing, these force or pressure sensors typically have low sensing resolution and are usually made of stiff materials, which is not favorable in scenarios that involve interaction with humans or delicate objects. On the other hand, vision-based tactile sensors have high sensing resolution and soft, skin-like surfaces. 
Therefore, this work uses vision-based tactile sensing in order to obtain rich tactile information and improve the dexterity of underactuated fingers in grasping and manipulation tasks. 

\subsection{Vision-based Tactile Sensing}

 Vision-based tactile sensor captures images of a gel surface with one or more miniaturized cameras, which provide high-resolution geometrical information about the contacted object.
 Moreover, varied tactile information can be distilled from tactile images, such as contact force, slip, etc \cite{wang2021gelsight, yuan2017gelsight, taylor2022gelslim, do2022densetact, sun2022soft}. 

Currently, there are many different vision-based sensors designed for parallel-jaw grippers and fingertips of multi-phalanx fingers \cite{wang2021gelsight, taylor2022gelslim, padmanabha2020omnitact, tippur2023gelsight360, lambeta2020digit, zhao2023gelsight}. However, these designs cannot be directly translated to fully sensorize multi-phalanx finger designs. There are several attempts to integrate tactile sensing with multi-phalanx fingers. Wilson \textit{et al.} proposed a two-finger gripper that has four phalanges and equipped each phalanx with one tactile sensor \cite{wilson2020design}. Similarly, Liu \textit{et al.} designed a two-phalanx soft finger with an embedded camera for each phalanx \cite{liu2023Endo}. This approach works well for hands with a small number of phalanges but poses challenges when a larger number of cameras are required for a humanoid hand. The need for a large number of video streaming and communication adds to the system's complexity.

Inspired by the use of mirrors to reduce finger size in GelSlim \cite{donlon2018gelslim} and Gelsight Wedge \cite{wang2021gelsight}, or to increase sensing area in GelSight Svelte \cite{zhao2023gelsight}, this work sensorizes a three-phalanx finger with only one camera and two mirrors. By optimizing the positions of the mirrors, the camera obtains a continuous view of the entire sensing surface for all possible finger configurations. Meanwhile, the camera perspective changes as the finger moves, from which the current finger configuration, i.e. proprioception, can be extracted.

\section{DESIGN AND FABRICATION}

The main design goal for GelLink was to ensure that the resulting finger is compact and low-cost, yet versatile.
Therefore, we allow only one motor and one camera per finger.
Furthermore, our design efforts are devoted to increasing torque transmission efficiency and tactile sensing coverage. 

\begin{figure}[!ht]
    \centering
    \includegraphics[width=0.48\textwidth]{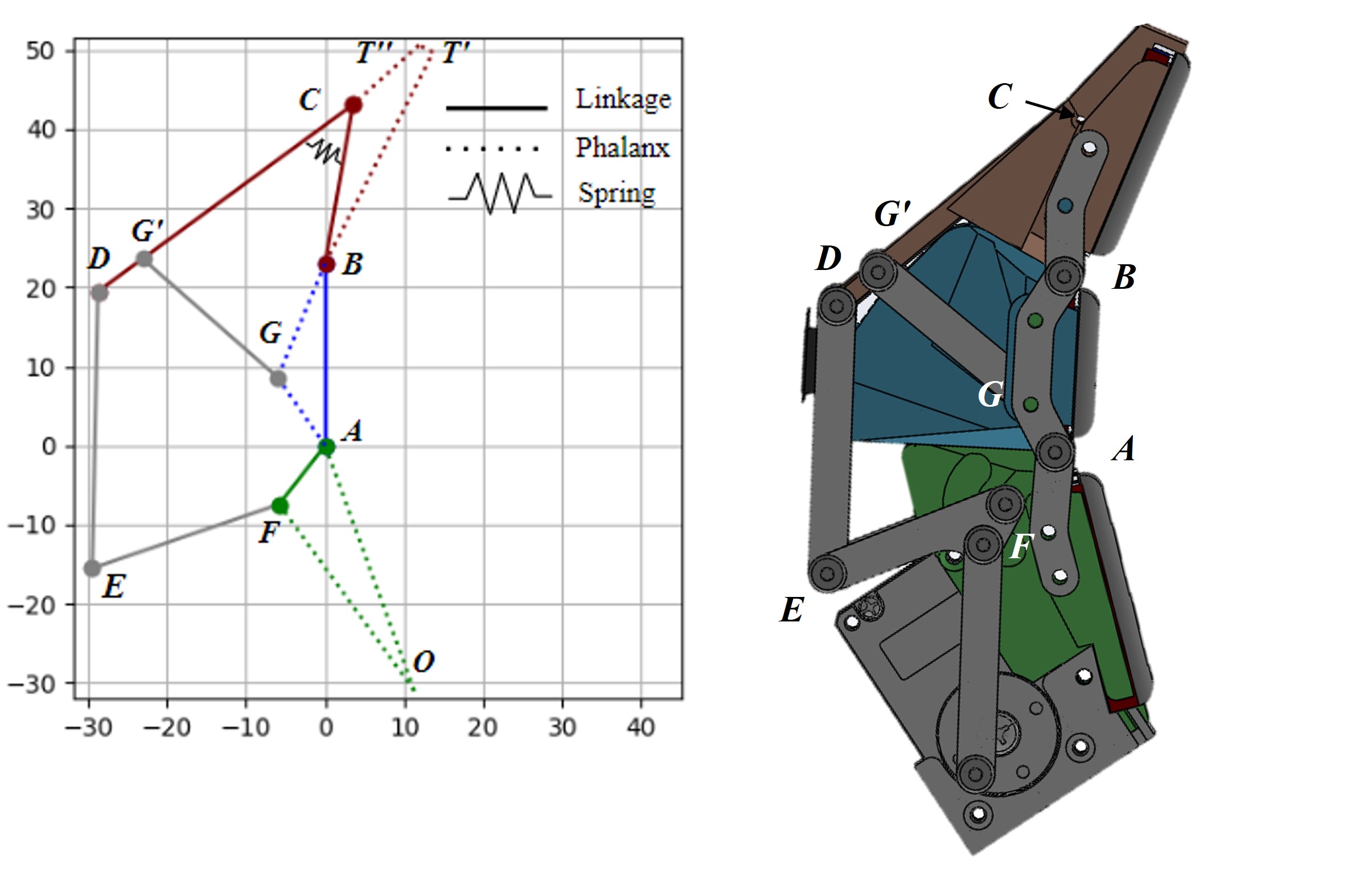}
    \caption{ \textit{Left:} Linkage model of GelLink. Solid lines represent the actual linkage system. Dashed lines represent the outlines of the phalanges, where red, blue and green lines correspond to distal, intermediate, and proximal phalanges respectively. The gray lines represent aluminum bars. \textit{Right:} Side view of the finger, where nodes are labeled in a similar manner to the linkage model.}
    \label{fig:flink}
\end{figure}

\begin{figure}[!ht]
    \centering
    \includegraphics[width=0.48\textwidth]{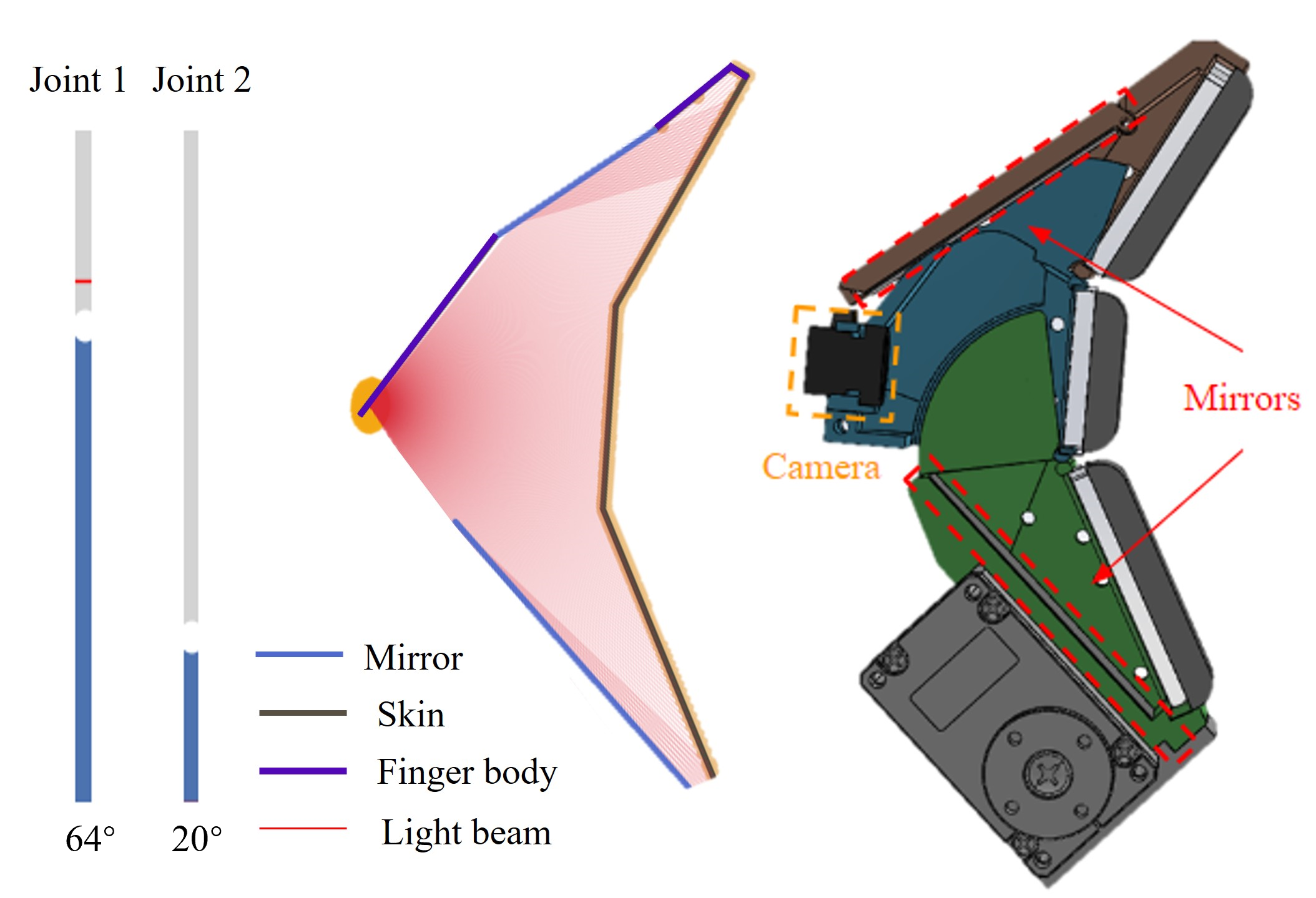}
    \caption{\textit{Left:} 2D reflection simulation of GelLink. Joint 1 is the proximal interphalangeal joint (PIP), and Joint 2 is the distal interphalangeal joint (DIP).  \textit{Right:} Section view of the finger.  }
    \label{fig:tsimu}
\end{figure}

As shown in Fig. \ref{fig:flink}, GelLink consists of three phalanges, two joints, one actuator, and one camera. 
The three phalanges are designed to mimic the distal, intermediate, and proximal phalanges of human fingers. 
The dimensions of these phalanges are determined based on anthropometric studies on human index fingers \cite{anthropometric}. 
The lengths of distal, intermediate, and proximal phalanges are \qty{32}{\mm}, \qty{23}{\mm}, and \qty{35}{\mm}, respectively. 
All three phalanges have a uniform width of \qty{34}{\mm}, which includes a sensing pad and transmission linkages on the sides.
Each sensing pad measures \qty{24}{\mm} wide, similar to human fingers' width. 
The \qty{10}{\mm} difference in width stems from the linkage and the shells of the phalanx bodies, ensuring essential structural strength.
The phalanges can be reduced in width to achieve a more compact design, albeit at the cost of reducing the sensing area.
Linkage mechanisms are situated on both sides of the finger, which efficiently transmits actuation torque, as shown by Fig. \ref{fig:flink}. The section view in Fig. \ref{fig:tsimu} illustrates the internal hollow structure of the phalanges, and offers a clear light path for camera-based tactile sensor integration. The mirrors installed on the distal and proximal phalanges allow the camera mounted on the intermediate phalanx to capture images of all sensing pads at all permissible finger configurations.

\subsection{Linkage Mechanism Design}
The proposed finger employs a 2D linkage mechanism to transmit torques with high efficiency, which is optimized using a 2D linkage simulator package \cite{mechanism}. 
The goal of the optimization is to improve the range of motion and retain a good transmission angle. We adopted a relaxed transmission angle limit of 35\degree-145\degree, because the amount of transmitted force is not very high and the mechanism runs at a low speed \cite{balli2002transmission, tao1964applied}.
We can acquire the positions, velocities, and accelerations of all linkages and pivot points using the kinematic analysis provided by the package.

\begin{figure}[h!]
    \centering
    \includegraphics[width=0.48\textwidth]{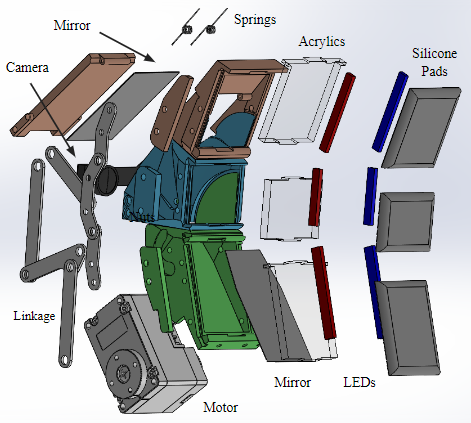}
    \caption{Exploded view of GelLink. Main bodies of distal, intermediate, and proximal phalanges are rendered in red, blue, and green, respectively.}
    \label{fig:exploded}
\end{figure}

Fig. \ref{fig:flink} illustrates the linkage structure of GelLink. Linkages and supporting structures of different phalanges are shown in their corresponding color, i.e. red for the distal phalanx, blue for the intermediate phalanx, and green for the proximal phalanx. The actual linkage mechanism consists of two parts: a six-bar linkage represented by the solid line loop $ABCDEFA$ and an extra constraint bar $GG'$. With Kutzbach's equation \cite{angeles2013rational}, the number of DOFs should be $3(L-1) - 2J - H$, where $L$ is the number of links, $J$ is the number of binary joints or lower pairs, and $H$ is the number of higher pairs. In this case, the mechanism has seven links and eight revolute joints, which lead to a two-DOF linkage mechanism. Link $EF$ is the actuation bar in the mechanism, which is indirectly actuated by a DYNAMIXEL motor through a four-bar parallel linkage. 
To add passive compliance to the distal interphalangeal joint, a spring is attached between links $BC$ and $CD$, as shown by Fig. \ref{fig:exploded}. $OA$, $AB$, and $BT'$ correspond to the sensorized surfaces of the proximal, intermediate, and distal phalanges.

The lengths and positions of links can be manually adjusted in the simulator. After several iterations in the simulator, the linkage can provide a 90-degree range of motion for both interphalangeal joints. Meanwhile, transmission angles between adjacent links are within the range of 37\degree-143\degree, which means that errors in finger motion and fluctuations in torque are low \cite{balli2002transmission}. 



\subsection{Tactile Sensing Integration}

In order to implement thorough and consistent tactile sensing, we built a 2D reflection simulator, which computes light rays and predicts the tactile sensing contact area.

The reflection simulator renders mechanical component interactions between mirrors, sensing pads, and light rays, which are highlighted with red in Fig. \ref{fig:tsimu}. The basic assumption is that the reflected ray and the incident ray have equal angles. In the simulator, angles of distal and interphalangeal joints can be adjusted in real-time. The camera can also be tilted in real time to obtain a better view of the sensing pads. Fig. \ref{fig:tsimu} shows that all three finger segments are visible during finger motion, with an optimized arrangement of a camera and mirrors.

\subsection{Fabrication}
An exploded view of the robotic finger is displayed in Fig. \ref{fig:exploded}. The finger fabrication can be divided into two parts: mechanical components and tactile sensing components. 

\begin{figure*}[h!]
    \vspace{2 mm}
    \centering
    \includegraphics[width=\textwidth]{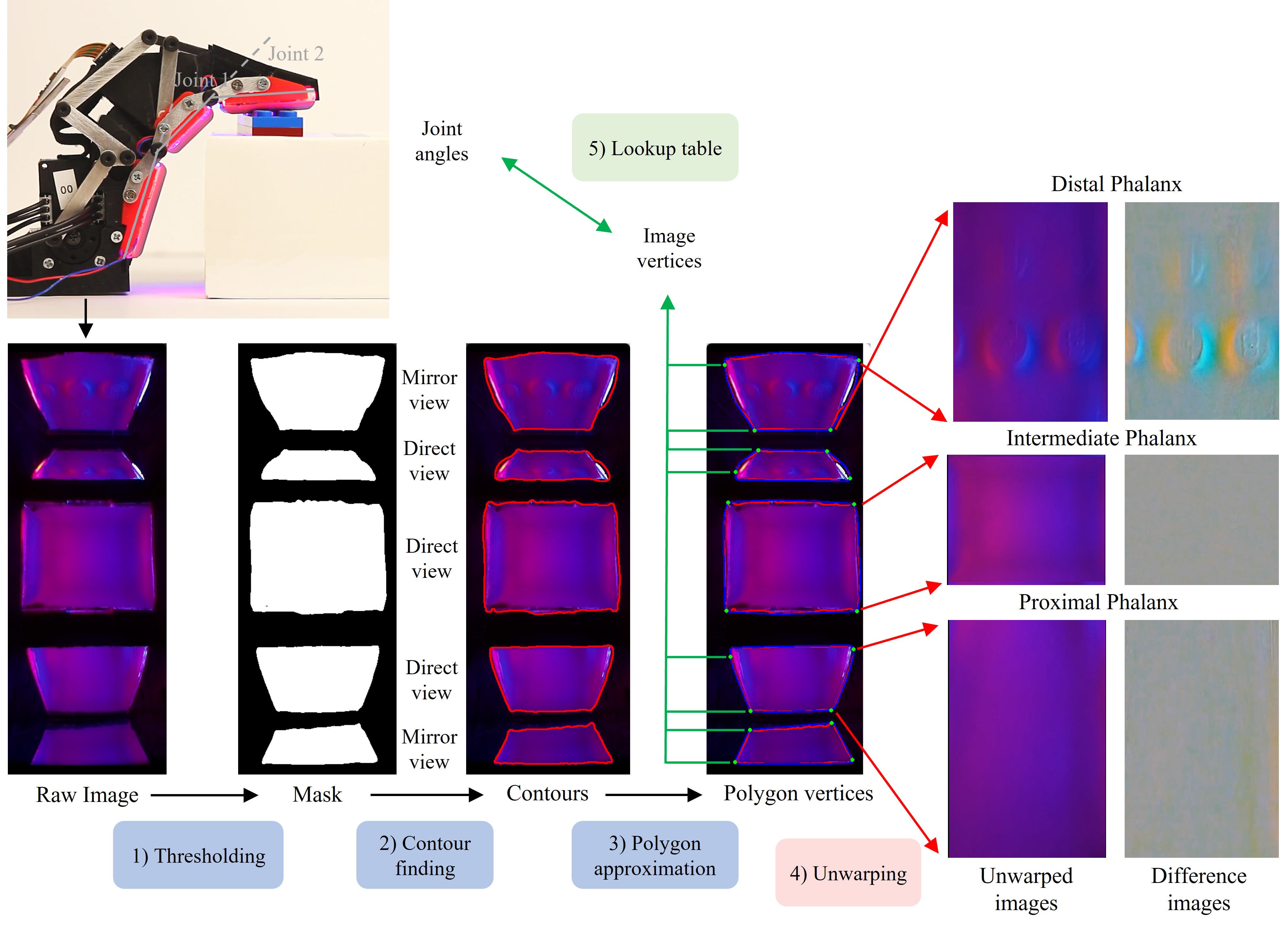}
    \caption{Workflow of image data interpretation. 1) Global thresholding is applied to the captured raw image to generate masks for distorted tactile images. 2) Contours and convex hulls can be easily found from the mask image, which are highlighted with red and blue lines in the figure. 3) Convex hulls are fed to polygon approximation algorithm \cite{ramer1972iterative}, which adapts contours with polygons and outputs the feature corners highlighted by green dots. 4) Unwarped tactile images and difference images (calculated as an after-contact tactile image subtracting an initial non-contact tactile image) can be computed. 5) A lookup table between joint angles and polygon vertices can be established for proprioceptive sensing.}
    \label{fig:workflow}
\end{figure*}

Mechanical components include phalanges, linkage bars, shoulder screws, lock nuts, and springs. The phalanges are 3D printed with PLA using a Prusa MK3S+ printer. The phalanges are carefully designed to house elastomer pads and the camera and allow the movement of links and joints. Thin fins on top of the intermediate and proximal phalanges are designed to block external illumination during finger movements. Tangentially, a piece of black stretchable cloth is attached between the intermediate and proximal phalanges. The linkage bars are machined from 6061 aluminum sheets with an OMAX waterjet machine. We used shoulder screws and lock nuts to allow low-friction rotation of linkage bars, which makes force transmission smooth and efficient. Torsion springs are integrated to add passive compliance and stability to the finger mechanism. 

Tactile sensing components consist of mirrors, acrylic pieces, silicone pads, a Raspberry Pi camera, and LED strips. 
The mirrors are laser cut from \qty{0.5}{\mm} PEG flexible mirror sheets, which can be easily attached to any flat surface. 
Clear acrylic pieces, which support the soft silicone pads, are also laser cut. 
The clear silicone pads are coated with a Lambertian gray paint, which provides accurate gradient information. The soft silicone is made from a mixture of XP-565 Parts A, B (Silicones Inc.), and a plasticizer (LC1550 Phenyl Trimethicone, Lotioncrafter) with a 1 to 12 to 3 parts ratio. 
The Lambertian paint is made of 1 part silicone ink catalyst to 10 parts gray silicone ink base to 1.25 parts $4 \mu m$ aluminum cornflakes to 30 parts NOVOCS Gloss (Raw Materials Inc., Schlenk, Smooth-on Inc), which is the same as the recipe used in \cite{liu2023gelsight}. The paint was evenly sprayed on the silicone pads with an airbrush. 
In order to achieve directional illumination with different colors, blue and red flexible COB LED strips (\qty{4}{\mm} COB LED strip light, OCONA) are attached to the sides of each acrylic piece, with color filters attached to prevent unwanted light radiation. 
A Raspberry Pi Zero Spy camera with a 160-degree FOV is used, and the video is streamed by mjpg streamer on the Raspberry Pi. With a Raspberry Pi 4B board, 30 FPS can be achieved for 1440$\times$1080 pixels. To get stable and well-blended color images, image gains are tuned for \textit{awbgainR} (red channel), \textit{awbgainB} (blue channel), and \textit{EV} (exposure compensation for RGB channels) with the \textit{AWB} (auto white balance) turned off \cite{wang2021gelsight}.

\section{RESULTS}
GelLink is able to measure contact geometry and its joint angles using one single camera. The image processing workflow is shown in Fig. \ref{fig:workflow}. Contact imprints and joint angles can be extracted from raw images. Finally, we conduct grasping experiments with GelLink, which can detect contact in tactile images and perceive object information, such as size and texture.

\begin{figure}[h!]
    \vspace{2 mm}
    \centering
    \includegraphics[width=0.48\textwidth]{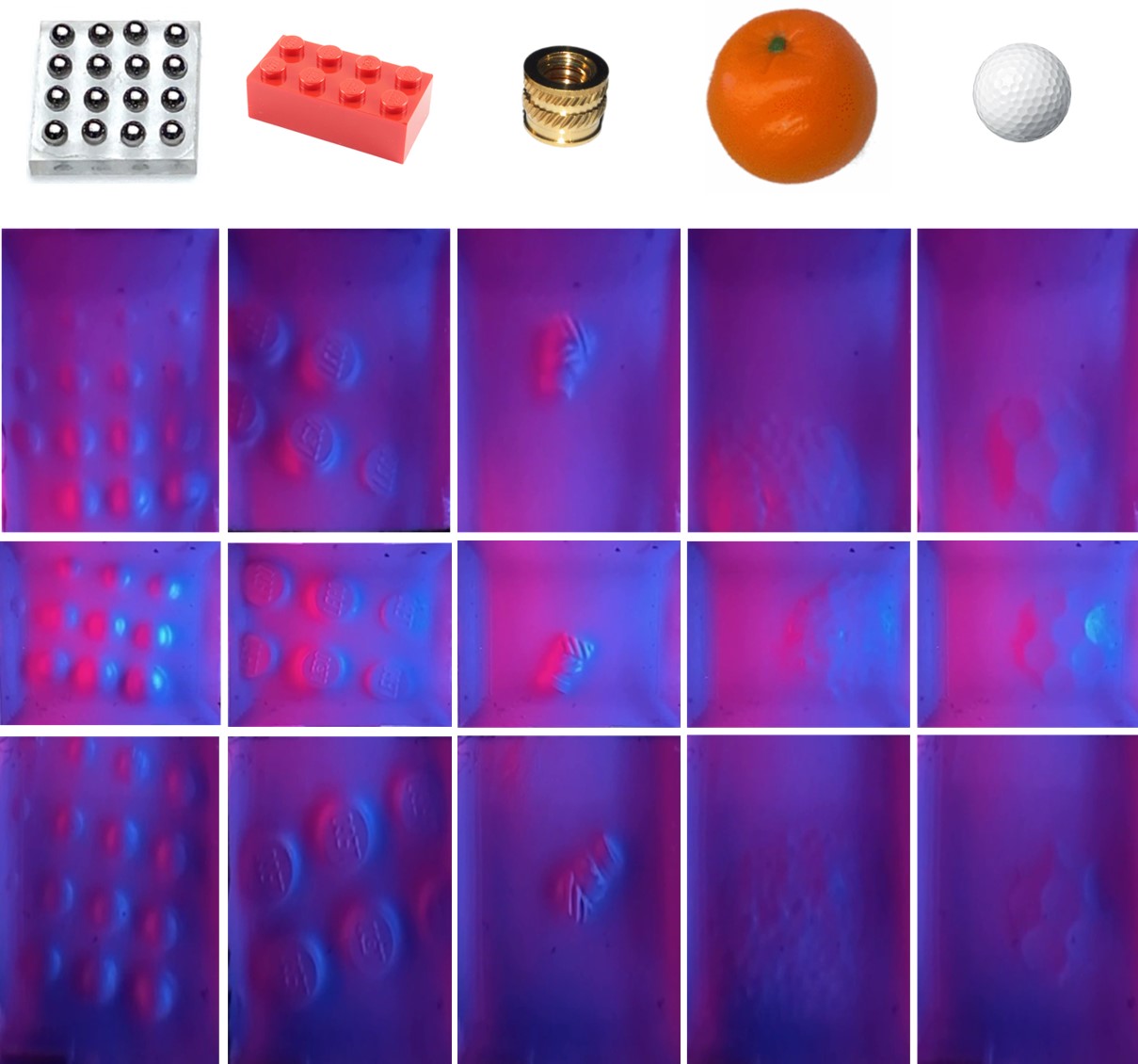}
    \caption{Tactile images of distal, intermediate, and proximal phalanges touching a matrix of calibration balls, a Lego piece, an M2.5 heat insert, a plastic orange, and a golf ball. From top to bottom: pictures of testing objects, where tactile images correspond to distal, intermediate, and proximal phalanges}
    \label{fig:contact}
\end{figure}

\subsection{Tactile Sensing}

To demonstrate the ability of continuous sensing, we used a calibration ball matrix, Lego piece, M2.5 heat insert, plastic orange, and golf ball to contact the different phalanges of GelLink. 
Fig. \ref{fig:contact} presents the tactile imprints from those experiments.
GelLink provides detailed information about grasped objects, which might benefit in-hand object classification, in-hand pose estimation, in-hand manipulation, etc.

\begin{figure}[h!]
    \vspace{2 mm}
    \centering
    \includegraphics[width=0.48\textwidth]{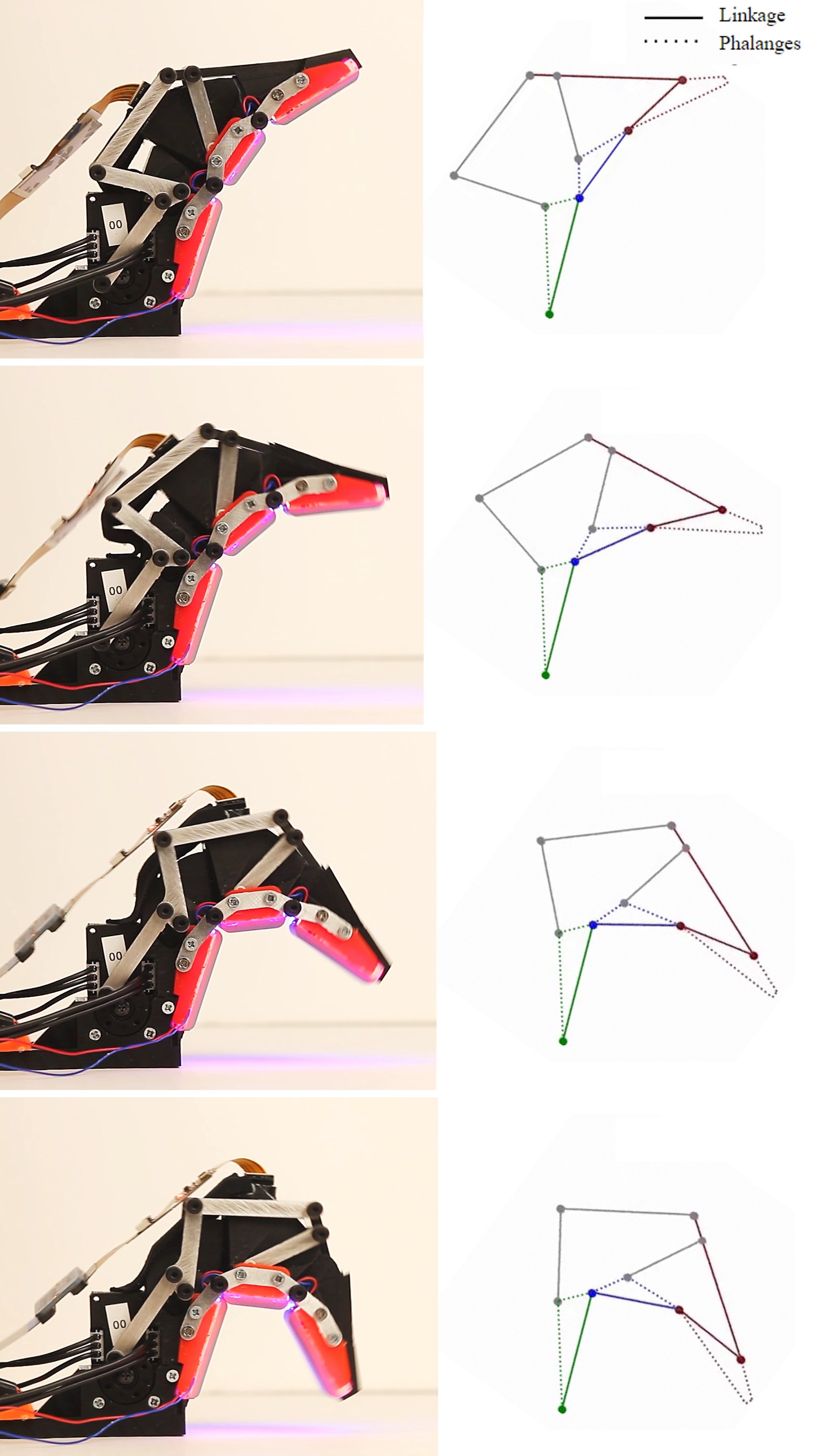}
    \caption{ Proprioception experiment. \textit{Left}: ground truth configuration of the finger and the corresponding tactile image. \textit{Right}: estimated finger configuration rendered with the linkage simulator}
    \label{fig:pro}
\end{figure}

\begin{figure}[h!]
    \centering
    \includegraphics[width=0.48\textwidth]{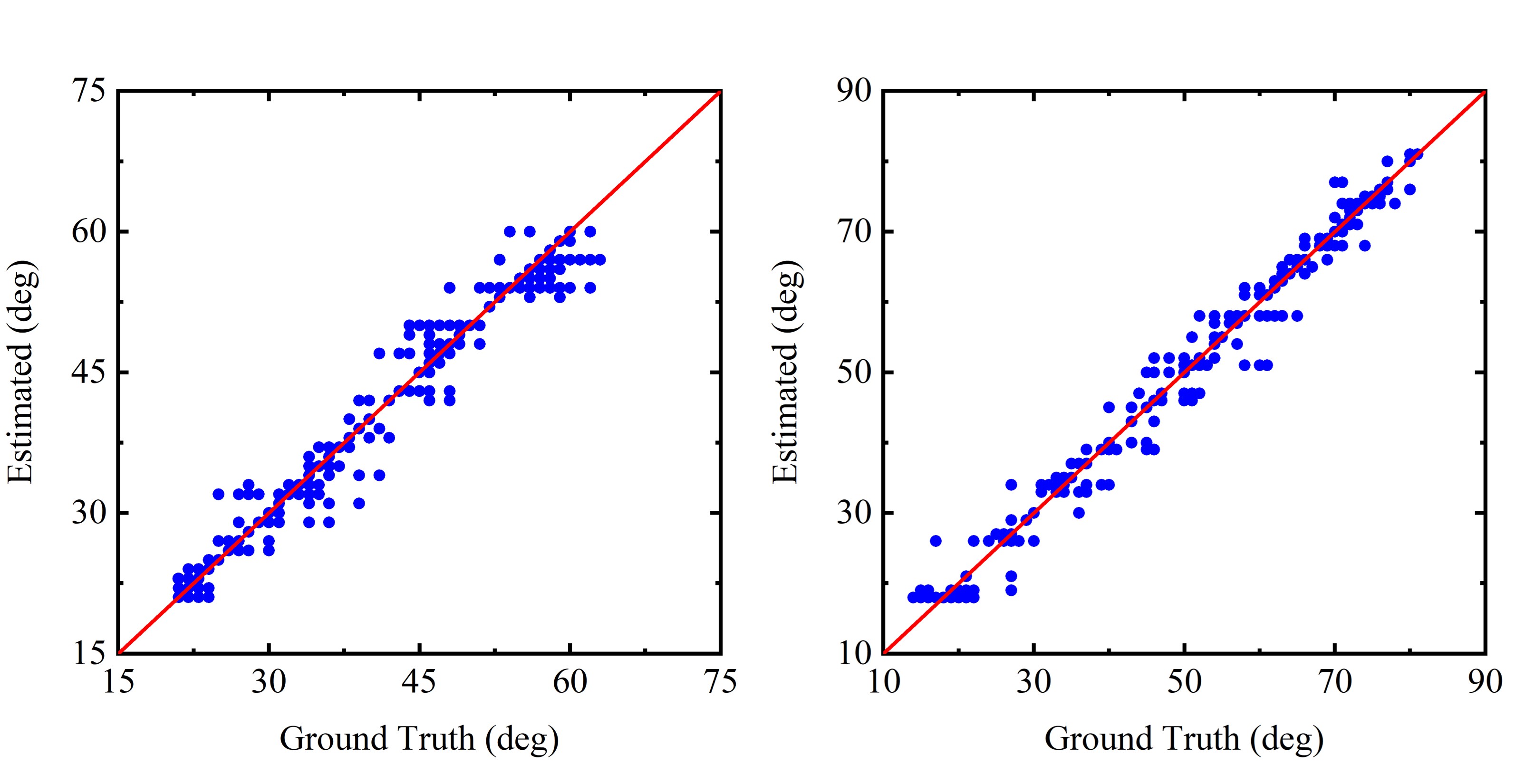}
    \caption{Distributions of estimated joint angles.}
    \label{fig:error}
\end{figure}

\subsection{Proprioception}
The relative positions of the camera, the mirrors, and the finger pads change as the finger moves, which further leads to changes in the raw images. 
We utilize such changes and extract proprioceptive information (i.e. joint angles) using the polygon vertices of the tactile images.

\begin{figure}[!h]
    \centering
    \vspace{2 mm}
    \includegraphics[width=0.49\textwidth]{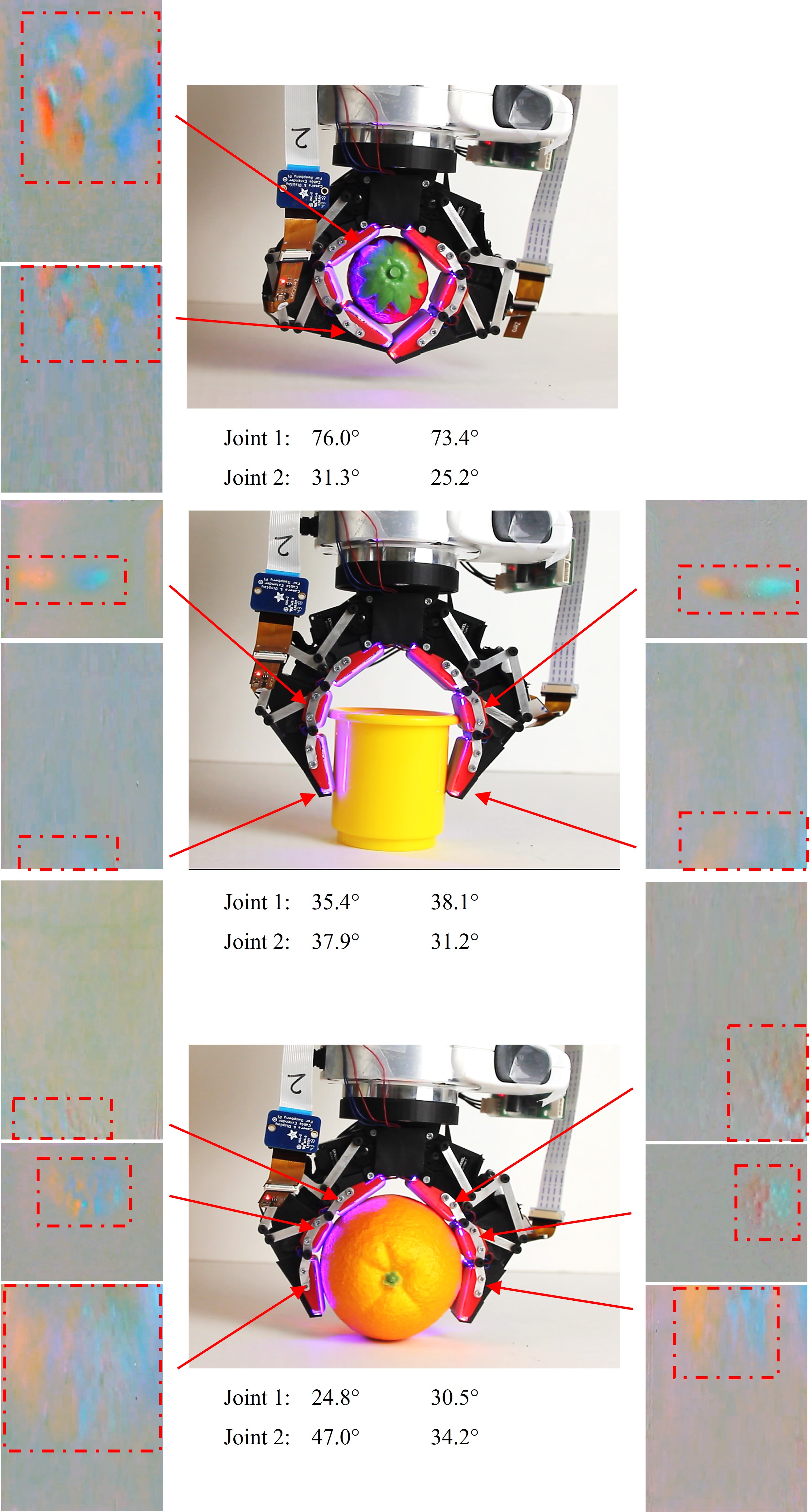}
    \caption{Object grasping experiment. We only display difference images when contacts are made between phalanges and objects. \textit{Top:} Grasping a plastic strawberry. \textit{Middle:} Grasping a toy cup. \textit{Bottom:} Grasping a plastic orange. }
    \label{fig:grasps}
\end{figure}

By recording and analyzing the finger flexing through its full range of motion, we establish a lookup table between joint angles and positions of polygon vertices. Then we evaluate joint angle estimation accuracy by repeating the same experiment. Ground truth values are acquired from the recorded video. The root mean square errors of the proximal interphalangeal joint and distal interphalangeal joint are 2.3\degree ~and 2.1\degree. Distributions of joint angle estimations are shown in Fig. \ref{fig:error}.

\subsection{Object Grasping}
GelLink can conform to objects with complicated shapes because of the spring-joint underactuation. Meanwhile, continuous tactile sensing and proprioceptive sensing abilities enable GelLink to detect contact and perceive information about grasped objects, such as texture and size. 
We conduct three object-grasping tasks using open-loop torque control and collect tactile and proprioceptive information from the grasps. The grasping was controlled with the current-based position control mode of the DYNAMIXEL motor, which closes the finger with a specified torque limit. The three objects are a plastic strawberry, a toy cup, and a plastic orange. 

We display tactile difference images captured from phalanges only where contacts are made. Red arrows are used to match tactile imprints with their corresponding contacts. Tactile imprints reveal texture and shape information about grasped objects. Joint angles estimated from the lookup table are also displayed in Fig. \ref{fig:grasps}. With the joint angles, we estimated the object widths to be \qty{48.8}{\mm} \qty{68.4}{\mm} and \qty{78.1}{\mm}, compared to their actual widths of \qty{45.2}{\mm}, \qty{70.0}{\mm}, and \qty{77.4}{\mm}. This tactile-based width perception could serve as a complement to visual perception, especially in situations where the gripper obstructs the view. 
The primary source of error stems from the imprecision in proprioception. The secondary factor is that our width estimation algorithm ignored the deformation of the contact pads.

\section{CONCLUSION AND DISCUSSION}
This paper presents GelLink, a novel robotic finger design that explores the symbiotic fusion of underactuation and vision-based tactile sensing.
We develop a planar linkage simulator and a planar reflection simulator to obtain a compact and low-cost finger design that requires less hardware but still has the advantages of adaptation and continuous sensing. 
Several experiments are designed to evaluate the quality of tactile sensing and proprioception and demonstrate the capability of grasping and perceiving.

The combination of underactuation and tactile perception allows GelLink to perform complex manipulation tasks that require intricate control policies and allows it to observe detailed object features. 
Moreover, the techniques developed in this paper will help researchers develop finger designs that combine low cost with good tactile and proprioceptive sensing, along with the ability to do underactuated grasping. Future work involves the incorporation of GelLink into a more dexterous hand, which will allow it to potentially perform more complex tasks.

\addtolength{\textheight}{-0cm}   


\section*{ACKNOWLEDGMENTS}

This work is financially supported by the Toyota Research Institute and Amazon Science Hub. We would like to acknowledge Sandra Q. Liu for her help in experiments and fabrication, and Megha H. Tippur for her help in designing PCBs. We would also thank Branden Romero, Maria Ramos Gonzalez, and Alan Papalia for their suggestions and feedback. 




\bibliographystyle{IEEEtran}
\bibliography{IEEEabrv,mybibfile}

\end{document}